\documentclass{article}
\usepackage{spconf,amsmath,graphicx}
\usepackage{graphicx}
\usepackage{amsmath}
\usepackage{amssymb}
\usepackage{booktabs}
\usepackage{multirow}
\usepackage{empheq}
\usepackage{subcaption}
\usepackage{arydshln}
\usepackage{lipsum}  

\usepackage{pifont}

\usepackage[pagebackref,breaklinks,colorlinks]{hyperref}

\usepackage[capitalize]{cleveref}
\crefname{section}{Sec.}{Secs.}
\Crefname{section}{Section}{Sections}
\Crefname{table}{Table}{Tables}
\crefname{table}{Tab.}{Tabs.}

\title{RESSCAL3D++: Joint acquisition and semantic segmentation of 3D point clouds}
\name{Remco Royen$^{\star}$, Kostas Pataridis$^{\dagger}$, Ward van der Tempel$^{\dagger}$, Adrian Munteanu$^{\star}$\thanks{This work is funded by Innoviris within the research project SPECTRE and by Research Foundation Flanders (FWO) within the research project G094122N.\\© 2024 IEEE. Personal use of this material is permitted. Permission from IEEE must be obtained for all other uses, in any current or future media, including reprinting/republishing this material for advertising or promotional purposes, creating new collective works, for resale or redistribution to servers or lists, or reuse of any copyrighted component of this work in other works. DOI: 10.1109/ICIP51287.2024.10647742}}
\address{$^{\star}$Department ETRO, Vrije Universiteit Brussel, Pleinlaan 2, B-1050 Brussels, Belgium\\
$^{\dagger}$VoxelSensors, Kantersteen 47, B-1000 Brussels, Belgium\\
Email: \{\textit{remco.royen, adrian.munteanu}\}@vub.be\\\{\textit{kostas.pataridis.ext, ward}\}@voxelsensors.be}

\begin{document}
%\ninept
%
\maketitle
\begin{abstract}

3D scene understanding is crucial for facilitating seamless interaction between digital devices and the physical world. Real-time capturing and processing of the 3D scene are essential for achieving this seamless integration. While existing approaches typically separate acquisition and processing for each frame, the advent of resolution-scalable 3D sensors offers an opportunity to overcome this paradigm and fully leverage the otherwise wasted acquisition time to initiate processing. In this study, we introduce VX-S3DIS, a novel point cloud dataset accurately simulating the behavior of a resolution-scalable 3D sensor. Additionally, we present RESSCAL3D++, an important improvement over our prior work, RESSCAL3D, by incorporating an update module and processing strategy. By applying our method to the new dataset, we practically demonstrate the potential of joint acquisition and semantic segmentation of 3D point clouds. Our resolution-scalable approach significantly reduces scalability costs from 2\% to just 0.2\% in mIoU while achieving impressive speed-ups of 15.6 to 63.9\% compared to the non-scalable baseline. Furthermore, our scalable approach enables early predictions, with the first one occurring after only 7\% of the total inference time of the baseline. The new VX-S3DIS dataset is available at \url{https://github.com/remcoroyen/vx-s3dis}.

% \begin{itemize}
%     \item Real-time point cloud processing is crucial
%     \item 
% \end{itemize}

\end{abstract}
%

% QUESTIONS:

\section{Introduction}
\label{sec:intro}

Given that human existence unfolds in a rich, three-dimensional physical realm, human perception relies on depth measurements. To allow our digital devices to interact in the same seamless manner with the physical world, 3D scanners play an important role. These devices facilitate the capture of 3D points in space, thereby providing a comprehensive and precise representation of the observed scene. Subsequently, achieving an accurate understanding of the acquired 3D scene is imperative, as it enables comprehension of the visual information and supports decision-making processes. To this end, a fundamental component is 3D semantic segmentation~\cite{qi2017pointnet++, zhao2021point, wu2022point, kolodiazhnyi2023oneformer3d}. This holds significant relevance in various domains such as robotics, autonomous driving, 3D modeling, extended reality, and other related applications. 

Various techniques, including LiDARs, scannerless Time-of-Flight (ToF), stereo vision, and structured light, are employed for 3D sensing. In the context of autonomous driving, LiDARs are commonly utilized~\cite{behley2019semantickitti, liao2022kitti}. While LiDARs offer high precision, they are expensive, bulky and demand significant power resources. In contrast, scannerless ToF sensors are more cost-effective but provide lower precision. Recently, VoxelSensors~\cite{voxelsensors} has introduced an innovative 3D scanning technology utilizing two cameras to track a laser that scans the 3D scene~\cite{van2023low}. This approach falls under the category of active stereo vision. The specific design of their sensor enables highly power-efficient and ultra-low-latency 3D sensing.

% Multiple techniques such as LiDARs, scannerless Time-of-Flight (ToF), stereo vision and structured light exist to achieve 3D sensing. In autonomous driving, LiDARs are often employed (\textcolor{red}{REFS}). While they achieve high precision, they are extremely expensive and have high power requirements. Scannerless ToF sensors are cheaper but achieve lower precision. Recently, VoxelSensors introduced a novel 3D scanners that employs 2 cameras to track a laser which scans the 3D scene (\textcolor{red}{REFS}). Thus, this can be categorized as an active stereo vision approach. The design of their sensor allows extremely power efficient and ultra-low-latency 3D sensing.

\begin{figure*}[!t]
    \centering
    \includegraphics[width=.98\textwidth]{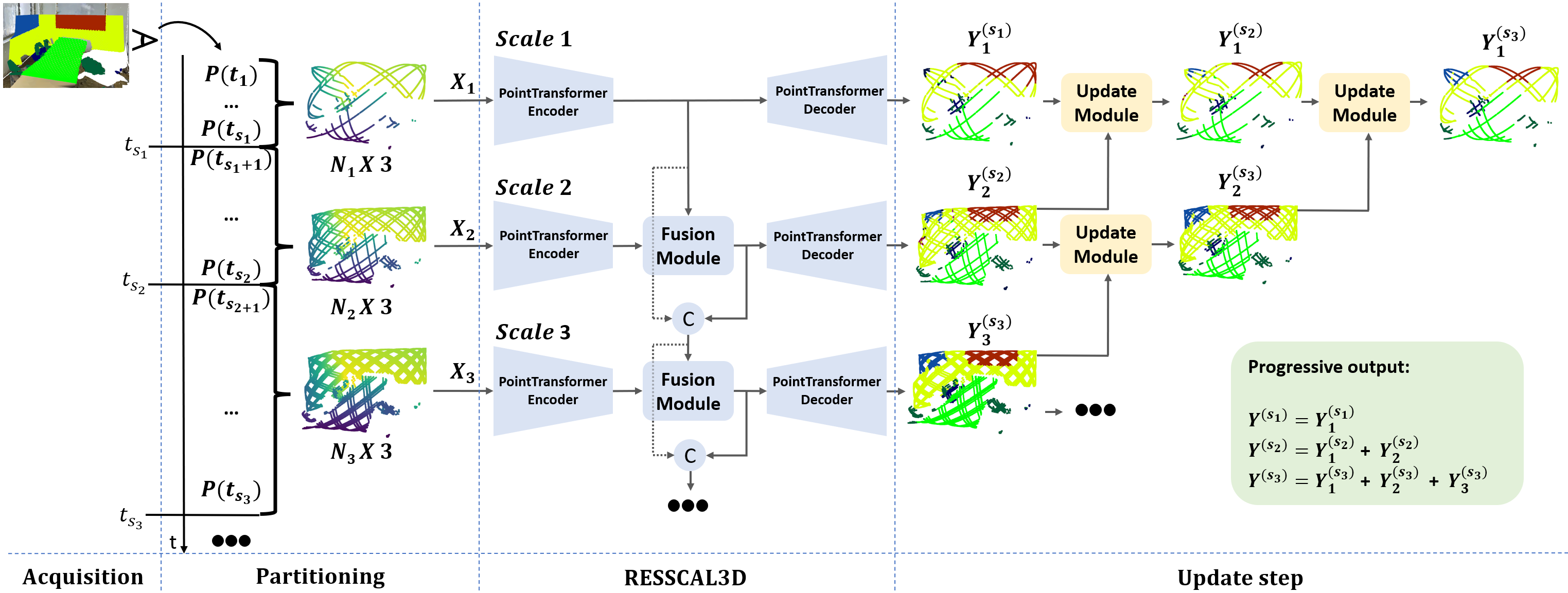}
    \caption{The RESSCAL3D++ architecture, enabling joint acquisition and processing.}
    \label{fig:arch}
\end{figure*}

VoxelSensors' 3D sensors offer an additional advantage through the use of a Lissajous scanning pattern. This pattern allows the scanner to rapidly obtain a spatially complete yet coarse point cloud. As the Lissajous pattern continues scanning, additional points are progressively acquired, resulting in a denser point cloud. This feature enables resolution-scalable 3D sensing. Scalability, well-established in traditional methods, has recently found application in deep learning through different flavors such as complexity~\cite{tan2020efficientdet, lin2017runtime} and quality~\cite{royen2021masklayer} scalability. The concept of resolution scalability is of great interest in compression techniques for both traditional~\cite{denis2010scalable, khalil2018scalable} and machine learning-based approaches~\cite{guarda2020deep, mei2021learning}.

% Complexity scalablility is achieved in \cite{tan2020efficientdet, lin2017runtime} by allowing the adjustment of model complexity during inference. MaskLayer~\cite{royen2021masklayer} presents a deep learning-based solution for quality-scalable predictions. The concept of resolution scalability is of great interest in compression techniques, spanning both traditional~\cite{denis2010scalable, taquet2012hierarchical, khalil2018scalable} and machine learning-based approaches~\cite{guarda2020deep, mei2021learning}.

% Another important advantage of the VoxelSensors' 3D sensors is the usage of a Lissajous scanning pattern. This allows the scanner to retrieve a spatially complete but coarse point cloud after a low amount of time. While the Lissajous pattern continues to scan the scene, additional points are obtained progressively and the obtained point cloud will become denser. As a consequence, it enables resolution-scalable 3D sensing. Scalability has many different flavors and has been an important feature in traditional methods \textcolor{red}{REFS}. Recently, scalability has also found its way in deep learning-based architectures. In (\textcolor{red}{REFS}), the model complexity can be chosen during inference. MaskLayer ~\cite{royen2021masklayer} presents a deep learning-based solution to obtain quality scalable predictions. Resolution scalability has been of great interest for compression, both for traditional \textcolor{red}{REFS} and machine learning based \textcolor{red}{REFS} techniques.

In numerous applications, real-time acquisition and processing are essential for enabling immediate interaction between the agent and the environment. An illustrative scenario is the exploration of an unfamiliar scene requiring obstacle avoidance. The agent's reaction time is defined as the sum of acquisition and processing time. Non-scalable methods can only be launched when the complete frame is available and thus do not employ the acquisition time. In our prior work, RESSCAL3D~\cite{royen2023resscal3d}, we introduced an architecture capable of handling resolution-scalable input data for semantic segmentation tasks. Leveraging this capability, the acquisition time is utilized to initiate the processing of the point cloud promptly, thereby reducing the overall reaction time. 

In this paper, we firstly present an important improvement which allows to reduce the cost of scalability from the 2.1\% in RESSCAL3D to 0.6\% at the highest scale. Secondly, while RESSCAL3D shows results on a subsampled dataset, we also evaluate on a novel dataset which emulates the behavior of the resolution scalable 3D sensor from VoxelSensors. The presented dataset, dubbed VX-S3DIS, is the first dataset leveraging resolution scalable point streams. We have devised an approach to jointly acquire and semantically segment a 3D scene. The proposed method is illustrated in Figure \ref{fig:arch}. Summarized, our main contributions are as follows:

% For many application, real-time acquisition and processing are vital as it allows the agent to immediately interact with the environment. An example is the exploration of an unknown scene where obstacles have to avoided. The reaction time of the agent can be defined as the sum of the acquisition and processing time. In our prior work, RESSCAL3D ~\cite{royen2023resscal3d}, we presented an architecture that is capable to handle resolution-scalable input data for semantic segmentation. By utilizing this property, the acquisition time is employed to already start the processing of the point cloud and thus to reduce In this paper, we firstly present an important improvement which allows to reduce the cost of scalability from \textcolor{red}{???}\% to \textcolor{red}{???}\% at the highest scale. Secondly, while RESSCAL3D shows results on a subsampled dataset, we present a novel dataset which captured scenes precisely simulating the behaviour of the resolution scalable 3D sensor from VoxelSensors. The presented dataset, dubbed VX-S3DIS, is the first dataset for a resolution-scalable 3D sensor. This dataset allowed us to devise an approach to jointly acquire and semantically segment a 3D scene and we present results on the new dataset. Our results show that ... . Summarized, our main contributions are as follows:

\begin{itemize}
    \item The introduction of an update step which allows to refine previous predictions, leading to a significant reduction in the cost of scalability from 2\% to 0.2\% on VX-S3DIS with a negligible effect on the inference time.
    \item The presentation of the first semantic segmentation dataset, dubbed VX-S3DIS, that mimics the working of a resolution-scalable 3D sensor, allowing intra-scan processing.
    \item Exhaustive experimentation of the proposed method on two datasets, showing the potential of the method for joint acquisition and processing on VX-S3DIS, with an important gain of 15.6-63.9\% in inference time, while keeping the cost of scalability as low as 0.2\%.
    % \item The validation of the proposed update step on both datasets, showing performance improvements of ... and ... \% respectively. Reducing cost of scalability to ... .
    % \item An extensive experimentation that shows the importance of the proposed update step and 
    % A practical application of the concepts on a simulator of a real-world 3D scanner, showcasting the potential of joint acquisition and processing
\end{itemize}

% \begin{itemize}
%     \item The first deep learning-based approach, to the best of our knowledge, that provides both resolution and quality scalable 3D semantic segmentation
%     \item An update module that improves predictions from lower resolution levels using newer, higher resolution predictions, leading to a significantly increased performance.
%     \item A practical application of the concepts on a simulator of a real-world 3D scanner, showcasting the potential of joint acquisition and processing
% \end{itemize}

The paper is structured as follows: Section \ref{sec:vx_s3dis} and \ref{sec:Proposed method} introduce the new point cloud dataset VX-S3DIS and the proposed method, respectively. In Section \ref{sec:experiments} the experimental results are presented. Finally, Section \ref{sec:conclusion} concludes this work.
\section{Proposed dataset: VX-S3DIS}
\label{sec:vx_s3dis}

\begin{figure*}
    \centering
    \includegraphics[width=.8\linewidth]{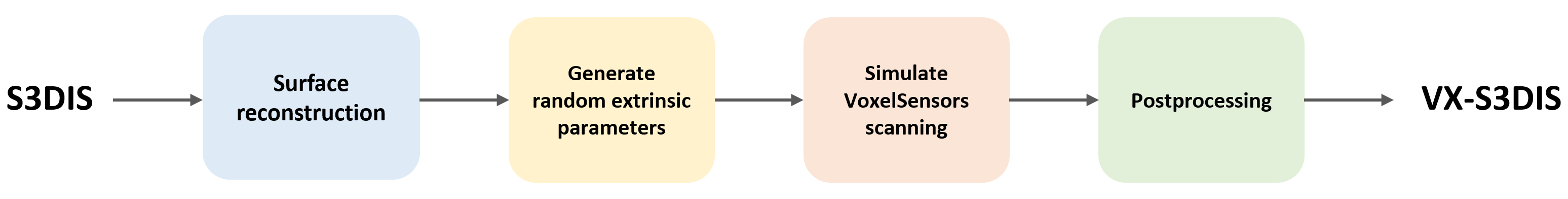}
    \caption{The data generation pipeline}
    \label{fig:dataset_pipeline}
\end{figure*}

\begin{figure*}[t]
\centering
\captionsetup{justification=centering}
\subcaptionbox*{(a)}{\includegraphics[width=0.24\textwidth]{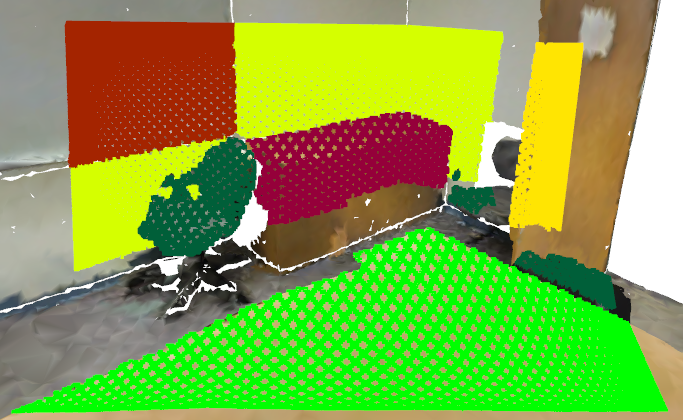}}
\subcaptionbox*{(b)}{\includegraphics[width=0.24\textwidth]{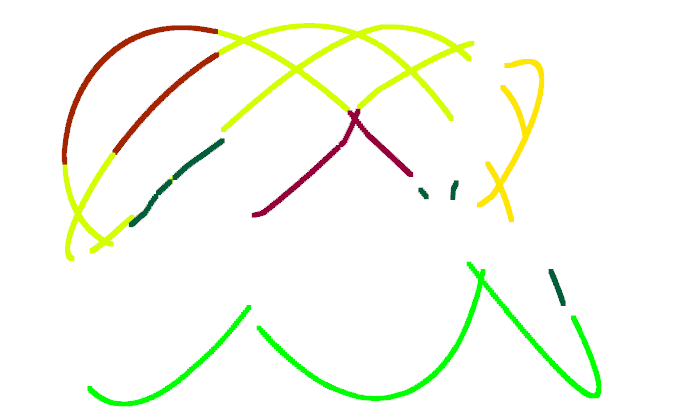}}
\subcaptionbox*{(c)}{\includegraphics[width=0.24\textwidth]{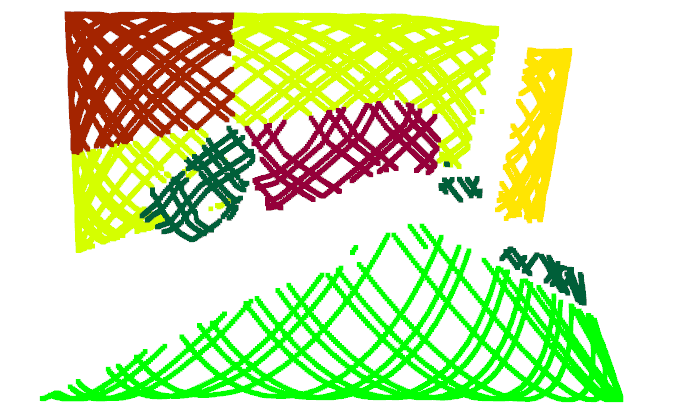}}
\subcaptionbox*{(d)}{\includegraphics[width=0.24\textwidth]{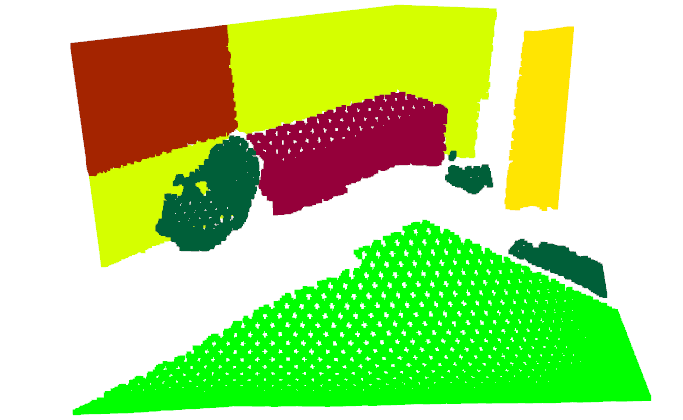}}
\captionsetup{justification=justified}
\caption{A sample of the VX-S3DIS dataset. (a) The sample visualised in the complete S3DIS room (b) the sample with all points until $t=2000$ (c) the sample until $t=15000$ (d) the full resolution sample, $t=65536$. The semantic labels were employed as colors to aide the visualization}
\label{fig:dataset_samples}
\end{figure*}

\textbf{Existing datasets.} The advent of deep learning has underscored the importance of accurately annotated datasets. For point cloud semantic segmentation, various datasets are employed for distinct applications. The Stanford 3D Indoor Scene dataset (S3DIS)~\cite{armeni20163d} and ScanNet~\cite{dai2017scannet} are popular datasets providing semantic annotations for indoor rooms, comprising dense point clouds for a large number of rooms. For autonomous driving, datasets such as SemanticKITTI~\cite{behley2019semantickitti} and KITTI-360~\cite{liao2022kitti} offer annotations for rotating LiDAR scans, capturing large scale complex traffic scenes.

% \begin{itemize}
%     \item S3DIS, Scannet: indoor rooms, popular for 3D semantic segmentation, whole rooms
%     \item kitty (semanticKitti and kitti-360) : autonomous driving: videos. Not resolution-scalable since the data is avilable only frame by frame and does not allow resolution scalable
% \end{itemize}

% No dataset exists that mimics the resolution scalable 3D scanning offered by recent 3D scanners.

% The Stanford 3D Indoor Scene dataset (S3DIS) \cite{armeni20163d} is a large-scale point cloud dataset that consists out of 271 indoor rooms. These rooms originate from 6 different areas where the 5-th area is located in a different building as the others. Therefore, Area-5 is often chosen as test dataset. Each point has rgb information and is annotated with one of the 13 semantic categories: ceiling, floor, wall, beam, column, window, door, table, chair, sofa, bookcase, board and clutter. S3DIS is a popular dataset and has often been used for the benchmark of semantic segmentation of point clouds. \textcolor{red}{An example sample can be seen in Figure ...} \textcolor{red}{Should we say something about existing datasets?}

\noindent\textbf{VX-S3DIS dataset} While the aforementioned datasets provide rich point clouds, they lack access to resolution-scalable sensor output. For instance, KITTI-360~\cite{liao2022kitti} offers sequential data in the form of scan-per-scan data with a timestamp per scan. Consequently, the scene can only be processed once the full resolution of a single scan becomes available. Presenting VX-S3DIS, we introduce the first point cloud dataset enabling intra-scan semantic processing. This is achieved by employing the principles of the VoxelSensors 3D sensor. As a laser scans the scene with a Lissajous pattern, the accumulated point cloud becomes denser over time. However, after a brief duration, a spatially complete but coarse representation is already obtained.

Our dataset originates from the S3DIS dataset~\cite{armeni20163d}, and its creation follows the pipeline outlined in Figure \ref{fig:dataset_pipeline}. Since S3DIS consists solely of vertices and the simulator rendering necessitates faces, we firstly perform surface reconstruction~\cite{bernardini1999ball}. The resulting meshes are simplified using Quadric Edge Collapse Decimation~\cite{garland1997surface} after removing possible holes based on the number of edges a boundary is composed of.

% As a second step, a large set of random extrinsic camera parameters are computed to capture each room from a diverse set of angles. To make sure that each sample contains sufficient of the room, the virtual cameras are grounded to a plane parallel and \textcolor{red}{10cm} inwards from the outer walls of the room. In order to automatise this process, the instance labels from S3DIS are employed to define the parallel planes. These are then discretised from which then randomly up to 50 virtual camera positions are selected. To make sure no objects block the camera, the cameras are put at least at 1.5m altitude. The camera rotation is computed such that it is oriented towards the center of the room. This information, together with the room meshes, are then employed by the VoxelSensors simulator to precisely simulate the scanning by their sensor. As an output, a stream of points is obtained for each sample. Each point does not only have the xyz coordinates and semantic labels but are ordered in the order they were captured and have a timestamp. In Figure \ref{fig:dataset_samples}, a dataset sample is visualised. Also different partitions based on timestamp are shown.

As a second step, a comprehensive set of random extrinsic camera parameters are computed to capture each room from diverse angles. To ensure each sample to contain sufficient room context, virtual cameras are grounded to a plane parallel and 5cm inwards from the outer walls. To automate this process, S3DIS instance labels are used to define the parallel planes, followed by a discretization. Up to 50 virtual camera positions are then randomly selected from the discretised locations. To prevent object obstruction, the cameras are positioned at a minimum altitude of 1.5m, with their rotation oriented towards the center of the room. This information, along with the room meshes, is utilized by the VoxelSensors simulator to precisely simulate the scanning by their sensor. The output consists of a stream of points for each data sample, in with each point consists of the xyz coordinates, a semantic label, and a timestamp. The order of the points in the stream is also the order in which the points were captured by the Lissajous pattern. Figure \ref{fig:dataset_samples} visualizes a dataset sample and example partitions based on the timestamps.

% The dataset can be represented as follows:
% \begin{equation}
%     \mathcal{D}=\{(\boldsymbol{x}_{j}, \boldsymbol{l}_{j}, \boldsymbol{t}_{j}\}_{j=1}^{|\mathcal{D}|},
% \end{equation}
% where $\boldsymbol{x}_{j}$, $\boldsymbol{l}_{j}$, $\boldsymbol{t}_{j}$ represent the 

% \begin{itemize}
%     \item Turn S3DIS in a mesh dataset by normal estimation, surface reconstruction, closing holes.
%     \item Compute random extrinsic camera parameters
%     \item Use VX simulator which emulates the working of the VX camera
%     \item Postprocessing to only retain xyz, label and timestamp
% \end{itemize}

\noindent\textbf{VX-S3DIS details.} The dataset comprises a total of 7031 samples derived from 168 distinct rooms within the S3DIS dataset. To ensure a sufficiently broad field-of-view, only offices, conference rooms, and auditoria are utilized from the original S3DIS dataset. The data is organized into different areas, mirroring the structure of the original dataset. The scanning pattern employs a Lissajous pattern with frequencies 1.1 and 1.8 mHz. The dataset includes annotations for 11 different classes: floor, wall, column, window, door, table, chair, sofa, bookcase, board, and clutter.

\section{Proposed method}
\label{sec:Proposed method}

\textbf{Overview of RESSCAL3D.} In our prior work, RESSCAL3D~\cite{royen2023resscal3d}, we introduced an architecture designed to handle resolution-scalable input data for semantic segmentation. In brief, it utilizes multiple branches that operate on different input resolutions. When low-resolution data becomes available, scale 1 can promptly generate an initial semantic prediction, even while higher resolution data is still being acquired. This allows for extremely early predictions and facilitates early decision-making. The subsequent branches process only the new, additional points and leverage features from previous scales in a fusion module to enhance performance. An important consequence of this approach is that RESSCAL3D processes all input points 31-61\% faster than the baseline method. While RESSCAL3D demonstrated a novel concept with compelling results, it exhibited two primary limitations, which we address in the enhanced version, RESSCAL3D++, depicted in Figure \ref{fig:arch}.

\noindent\textbf{Application on 3D sensor data.} RESSCAL3D subsampled complete rooms from S3DIS into different resolution scales to demonstrate its functioning. While this serves as a proof of concept, it does not accurately reflect the working of a resolution-scalable 3D sensor. With the introduction of our VX-S3DIS dataset (Section \ref{sec:vx_s3dis}), we can devise a method to handle the resolution-scalable input stream of points and assess its performance on VX-S3DIS. We can express the stream of points $\mathcal{P}$ mathematically as follows:
% While RESSCAL3D demonstrated compelling results on uniformly subsampled complete rooms as partitions, it lacked the application on a point stream from a resolution-scalable 3D sensor. With the introduction of the first point cloud dataset presenting such data in Section \ref{sec:vx_s3dis}, we can devise a method to handle this resolution-scalable input stream of points and evaluate its performance on VX-S3DIS. We can express the stream of points $\mathcal{P}$ mathematically as follows:
\begin{equation}
\label{eq:stream_points}
    \mathcal{P} = \{P(t_1), \ldots, P(t_{s_1}), P(t_{s_1+1}), \ldots, P(t_{s_2}), \ldots \},
\end{equation}
where $P(t_i)$ is the point obtained at timestamp $t_i$. We divide $\mathcal{P}$ into multiple partitions based on the timestamps. We define $X_i \in \mathbb{R}^{N_{i} \times 3}$ as the $i$-th partition used as input for scale $i$ with $N_i$ points, with $N_{i-1}<N_i<N_{i+1}$. We can express $X_i$ as $\{P(t_{s_{i-1}+1}), \ldots, P(t_{s_{i}})\}$ with $t_{s_{i}}$ denoting the end of the partition for scale $i$. Importantly, each partition may not have a fixed cardinality, as it is possible that no point is detected at a specific time instance. Partitioning based on timestamps, rather than the number of points, ensures that each partition represents a specific resolution rather than a specific size. The partitioning of one data sample can be visualized in Figure \ref{fig:arch}. This approach enables the asynchronous launch of different scales as soon as a partition is acquired, effectively utilizing the acquisition time of the remaining scales to initiate processing.

% Secondly, since no dataset exhibiting resolution-scalable input data existed, RESSCAL3D was shown on an in different resolutions subsampled complete room. While it succeeded in showing the potential, it is not a practical application. With the introduction of VX-S3DIS in this paper, we are able to apply the concepts on a practical example to show its potential for joint acquisition and semantic segmentation. To do so, an approach has to be devised to handle the output stream from the 3D sensor. We propose to partition the incoming point cloud stream in different partitions of different sizes and thus resolutions. Important is that this happens based on the timestamp and not the number of points. This will ensure the same resolution to be captured over different samples for  , independent of possible non-detected points. This can be mathematically expressed as:
% MATHEMATIC
% with ... . The different partitions can be visualized in Figure \textcolor{red}{Ref to fig}. This approach allows to launch asynchronously the different branches as soon as a partition is complete and thus effectively employ the acquisition time to start the processing. $Y_i^{(s_i)}$

\noindent\textbf{Update Module.} In RESSCAL3D, each scale $i$ predicts the semantic labels $Y_i$, where subscript $i$ indicates predictions for input data $X_i$. While higher scales benefit from prior information through the fusion module, lower scale labels are not refined when more accurate predictions become available, leading to inconsistencies when accumulated. Thus, in RESSCAL3D++, we introduce an update module and strategy, depicted in Figure \ref{fig:arch}, to refine earlier predictions when new information emerges. Let $Y_i^{(s_i)} \in \mathbb{R}^{N_{i}}$ be the predictions obtained by scale $i$, where we added superscript $(s_{i})$ denoting all predictions up to scale $i$ were employed. When predictions from scale $i+1$, $Y_{i+1}^{(s_{i+1})}$, become available, this is employed by the update module to update $Y_{i}^{(s_{i})}$ to $Y_{i}^{(s_{i+1})}$. Similarly, When the output of scale $i+2$, $Y_{i+2}^{(s_{i+2})}$, becomes available, this allows to refine $Y_{i+1}^{(s_{i+1})}$ to $Y_{i+1}^{(s_{i+2})}$, which in turn further refines $Y_{i}^{(s_{i+1})}$ to $Y_{i}^{(s_{i+2})}$. This can be mathematically expressed as follows:
\begin{equation}
\begin{split}
\label{eq:update_module}
    Y_i^{(s_{i+2})} & = \text{UM}(Y_i^{(s_{i+1})}, Y_{i+1}^{(s_{i+2})})\\
    & = \text{UM}(\text{UM}(Y_{i}^{(s_{i})}, Y_{i+1}^{(s_{i+1})}), \text{UM}(Y_{i+1}^{(s_{i+1})}, Y_{i+2}^{(s_{i+2})})),
\end{split}
\end{equation}
where UM represents the update module. This module creates a neighborhood of the K-nearest neighbors from the subsequent scale $i+1$ for each point in scale $i$. A majority voting among these neighbors allows to obtain the refined predictions. The complete output at each scale $i$, $Y^{(s_{i})}$, is the concatenation of its own output with those of preceding scales, refined with all available information at $t_{s_{i}}$,
\begin{equation}
\begin{split}
\label{eq:complete_output}
    Y^{(s_{i})} & = \{Y_1^{(s_{i})}, Y_2^{(s_{i})}, \ldots, Y_{i-1}^{(s_{i})}, Y_{i}^{(s_{i})}\}.
\end{split}
\end{equation}

\begin{figure}[!t]
    \centering
    \includegraphics[width=.75\linewidth]{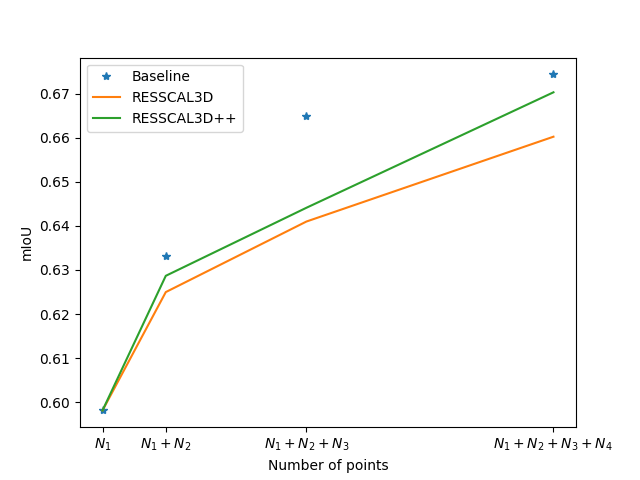}
    \caption{Comparison of RESSCAL3D++ with the non-scalable baseline~\cite{zhao2021point} and scalable state-of-the-art~\cite{royen2023resscal3d} on S3DIS}%Comparison in mIoU of RESSCAL3D with the non-scalable baseline and the ablation of the Fusion Module.}
    \label{fig:s3dis_mIoU_results}
\end{figure}

\begin{figure}[!t]
    \centering
    \includegraphics[width=.75\linewidth]{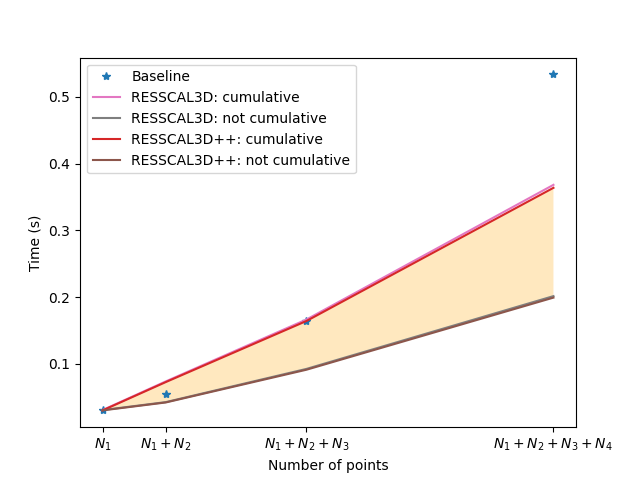}
    \caption{Comparison of RESSCAL3D++ with the non-scalable baseline~\cite{zhao2021point} and scalable state-of-the-art~\cite{royen2023resscal3d} in inference time on S3DIS. The actual inference latency is bounded to the yellow zone. The displayed non-scalable baseline timing results are not cumulative.}
    \label{fig:s3dis_time_results}
\end{figure}

% A first limitation of RESSCAL3D is the single-direction information flow. As the feature. While this  that eventhough the limited quality of predictions at the lower scales

% A first limitation of RESSCAL3D is that even though higher accuracies were obtained at higher scales, the lower 
% A first limitation of RESSCAL3D is the limited quality of predictions at the lower scales, obtained by processing the lower resolution input data. While the fusion module allows to improve performance in the higher scales, the lower scales remain unaltered. In RESSCAL3D++, we propose the introduction of an update module that . Since these predictions are obtained by processing lower resolution input data, they achieve a lower accuracy. 

\begin{figure*}[t]
\centering
\captionsetup{justification=centering}
{\includegraphics[width=0.24\textwidth]{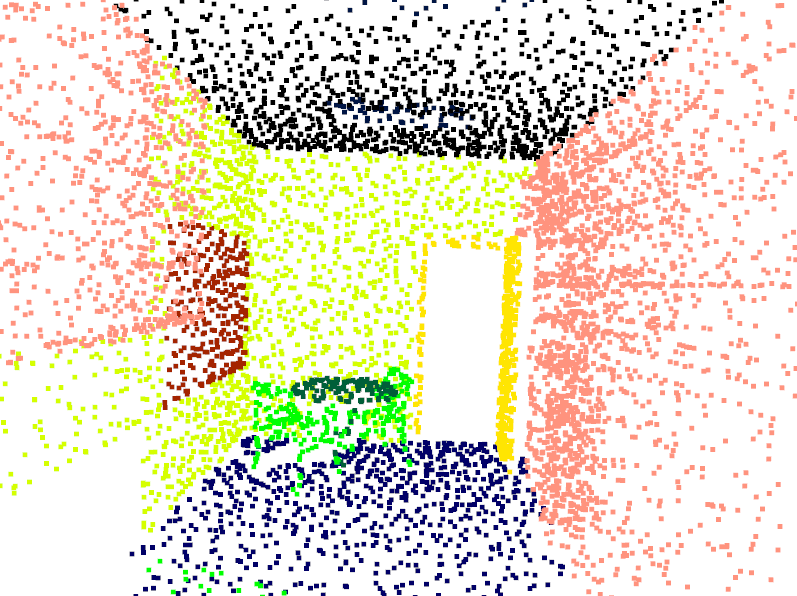}}
{\includegraphics[width=0.24\textwidth]{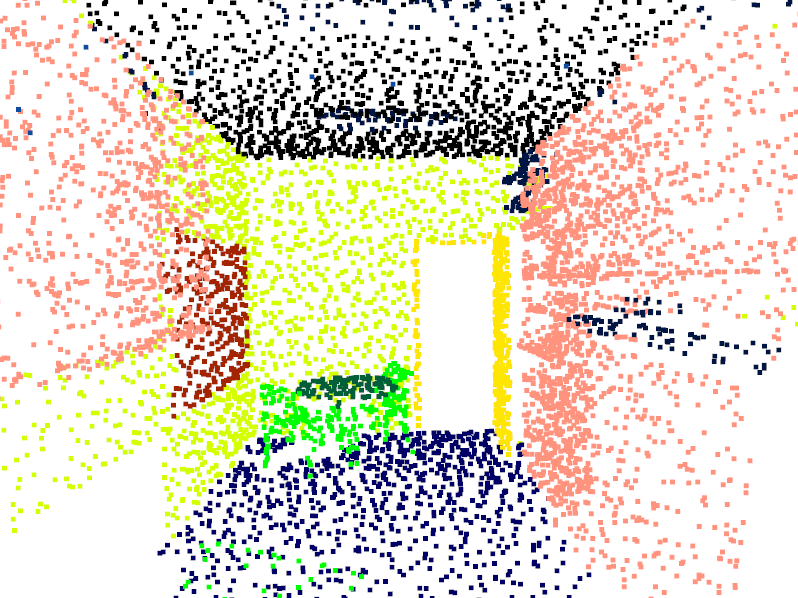}}
{\includegraphics[width=0.24\textwidth]{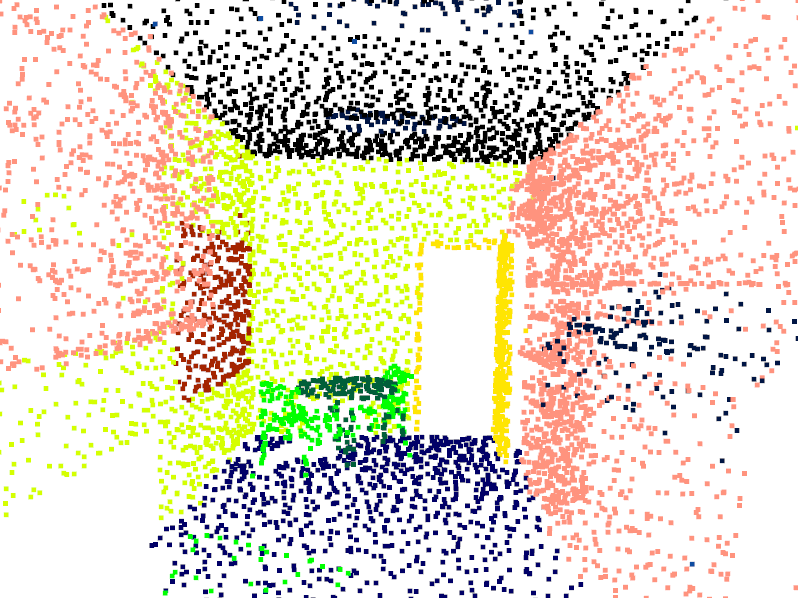}}
{\includegraphics[width=0.24\textwidth]{figs/visuals/new_visuals/s3dis_resscal3d_scale01.png}}
{\includegraphics[width=0.24\textwidth]{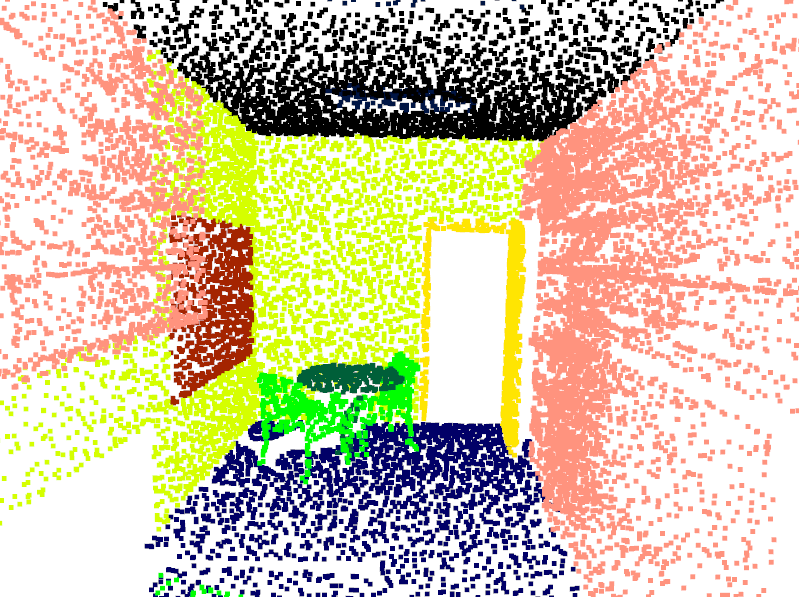}}
{\includegraphics[width=0.24\textwidth]{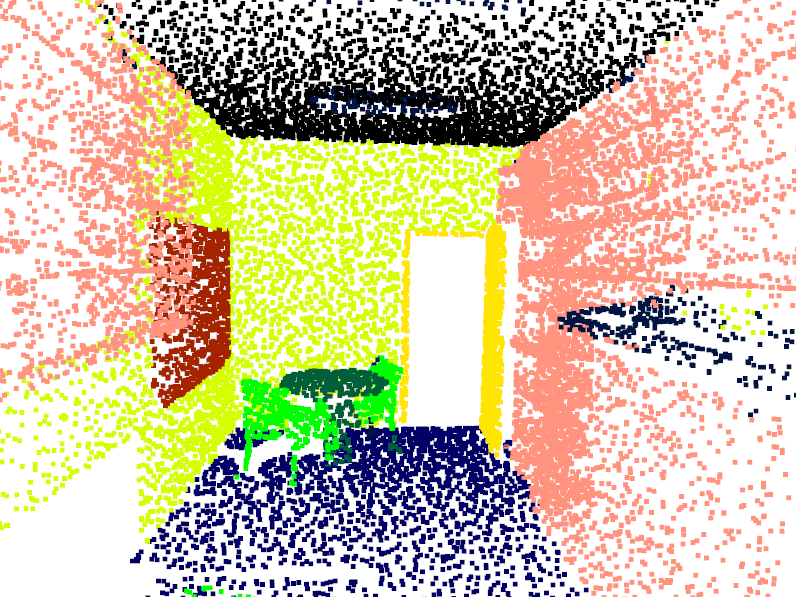}}
{\includegraphics[width=0.24\textwidth]{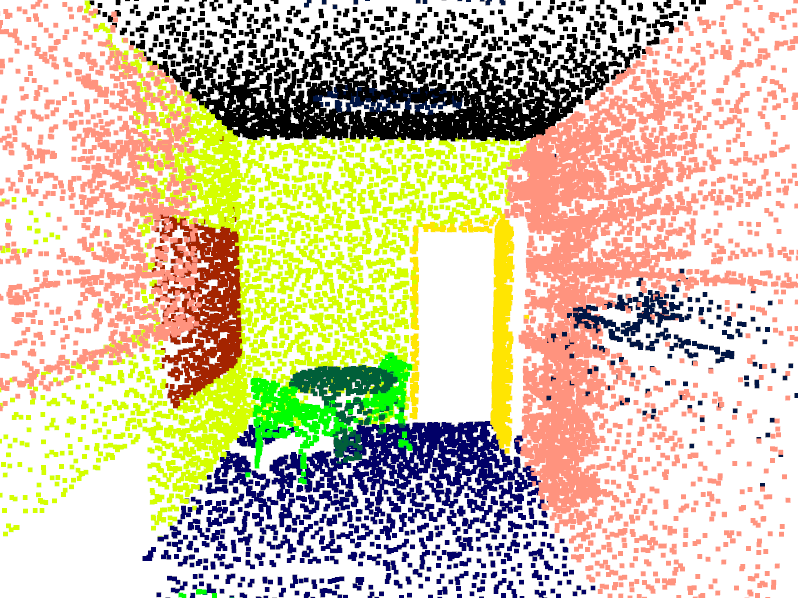}}
{\includegraphics[width=0.24\textwidth]{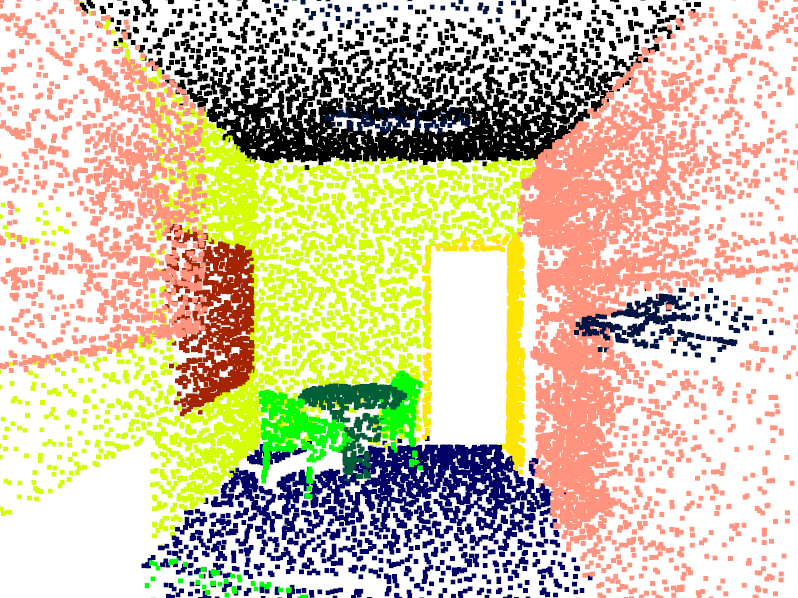}}
% \subcaptionbox*{(d)\\Scale 2\\Time: \textcolor{red}{???} s}{\includegraphics[width=0.19\textwidth]{figs/arch/resscal_fusion_arch.png}} 
\subcaptionbox*{(a)\\Ground truth}{\includegraphics[width=0.24\textwidth]{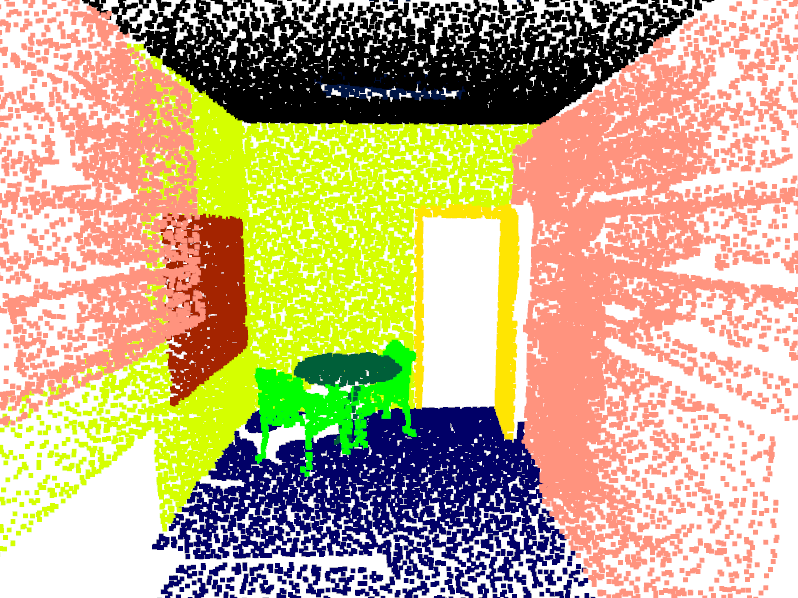}}
\subcaptionbox*{(b)\\Non-scalable baseline~\cite{zhao2021point}}{\includegraphics[width=0.24\textwidth]{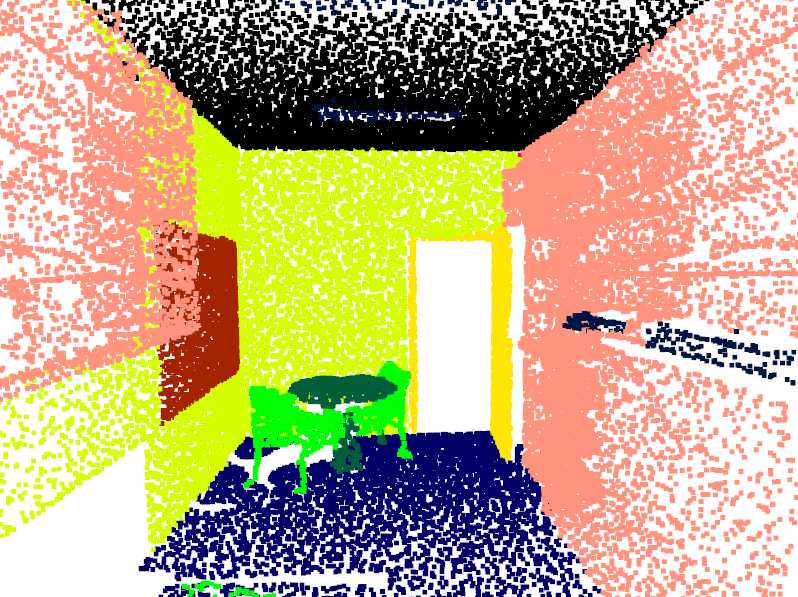}}
\subcaptionbox*{(c)\\RESSCAL3D~\cite{royen2023resscal3d}}{\includegraphics[width=0.24\textwidth]{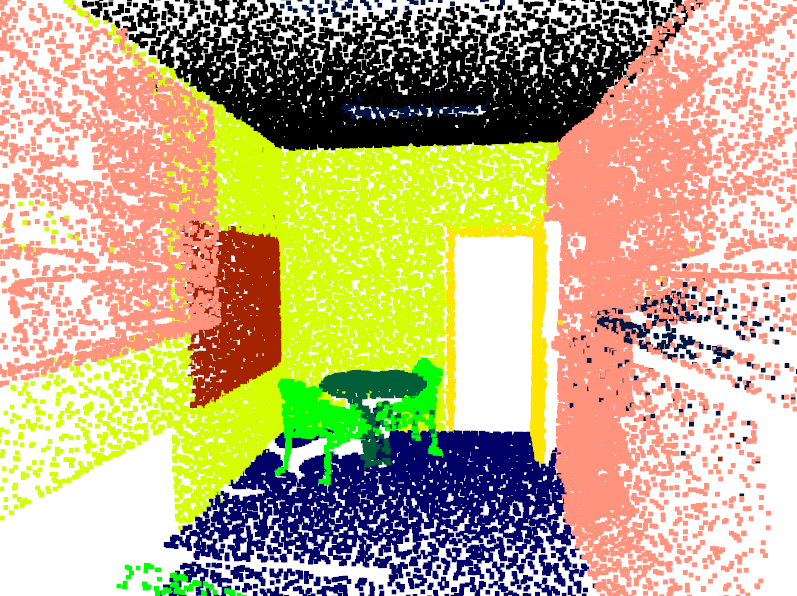}}
\subcaptionbox*{(d) \\RESSCAL3D++}{\includegraphics[width=0.24\textwidth]{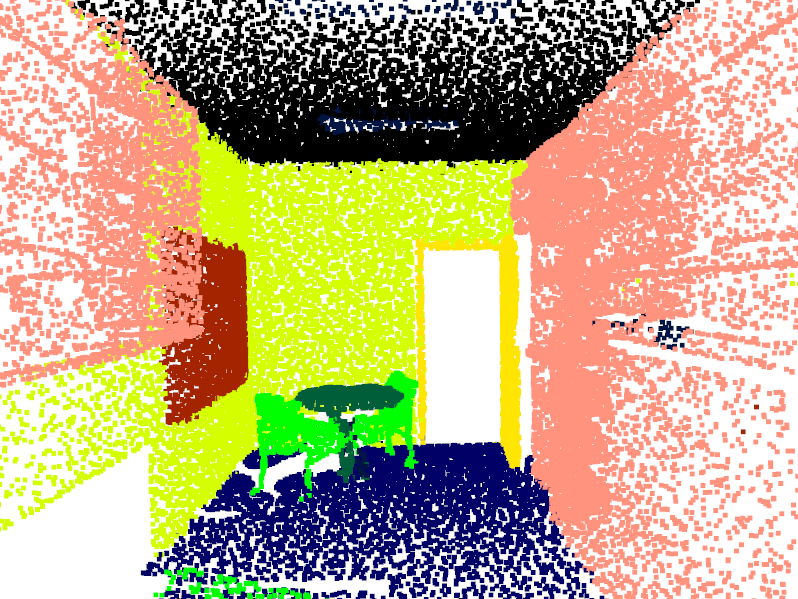}}
\captionsetup{justification=justified}
\caption{Visualization of the non-scalable baseline~\cite{zhao2021point}, RESSCAL3D~\cite{royen2023resscal3d} and RESSCAL3D++ on S3DIS. The semantic predictions after the second, third and fourth (last) scale are visualised on the top, middle and bottom row, respectively.}
\label{fig:s3dis_visuals}
\end{figure*}

% $Y(t_{s_{2}})$, $Y(t_{s_{3}})$ and $Y(t_{s_{5}})$

\section{Experiments}
\label{sec:experiments}

\begin{figure}[!b]
    \centering
    \includegraphics[width=.75\linewidth]{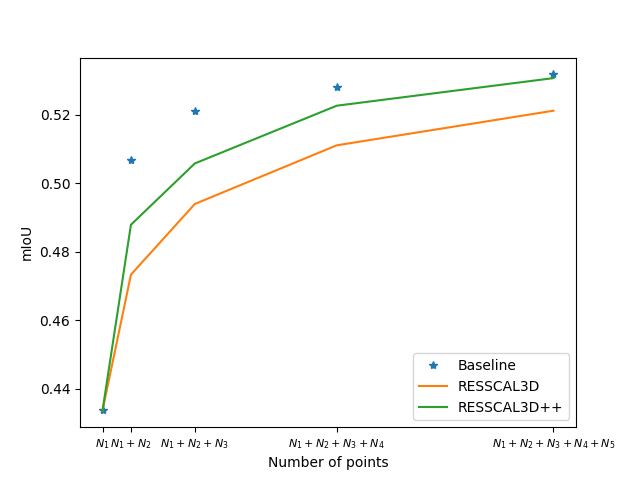}
    \caption{Comparison of RESSCAL3D++ with the non-scalable baseline~\cite{zhao2021point} and scalable state-of-the-art~\cite{royen2023resscal3d} on VX-S3DIS.}
    \label{fig:vx_miou_results}
\end{figure}

\begin{figure}[!b]
    \centering
    \includegraphics[width=.75\linewidth]{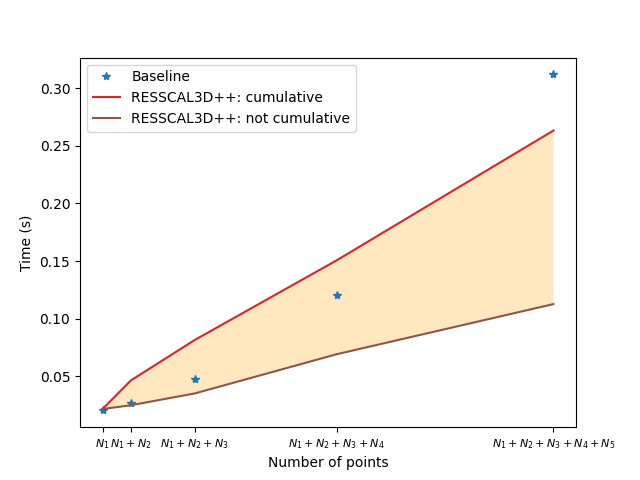}
    \caption{Comparison of RESSCAL3D++ with the non-scalable baseline~\cite{zhao2021point} and scalable state-of-the-art~\cite{royen2023resscal3d} in inference time on VX-S3DIS. The actual inference latency is bounded to the yellow zone. The displayed non-scalable baseline timing results are not cumulative.}
    \label{fig:vx_time_results}
\end{figure}

% \subsection{Dataset and Evaluation metrics}
\textbf{Datasets and Evaluation metrics.} To allow an easy comparison with RESSCAL3D~\cite{royen2023resscal3d}, we also apply RESSCAL3D++ on the S3DIS dataset \cite{armeni20163d}. To obtain different resolution levels, we follow \cite{royen2023resscal3d} and downsample S3DIS by voxelization with different voxel sizes. Secondly, we also apply the proposed method on the VX-S3DIS dataset, introduced in Chapter \ref{sec:vx_s3dis}. As evaluation metric, the mean intersection over union (mIoU) is being used as in \cite{zhao2021point, royen2023resscal3d}. All presented results are averaged over the Area-5 testset.

\begin{figure*}[t]
\centering
\captionsetup{justification=centering}
% Ground-truth
{\includegraphics[width=0.24\textwidth]{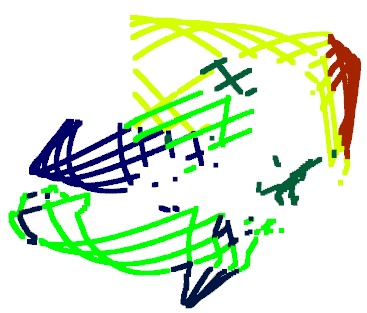}}
{\includegraphics[width=0.24\textwidth]{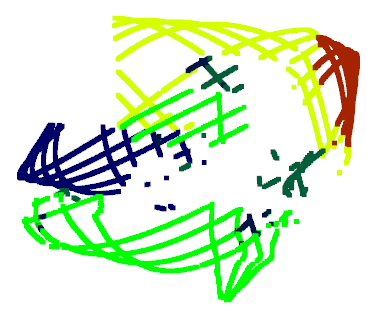}}
{\includegraphics[width=0.24\textwidth]{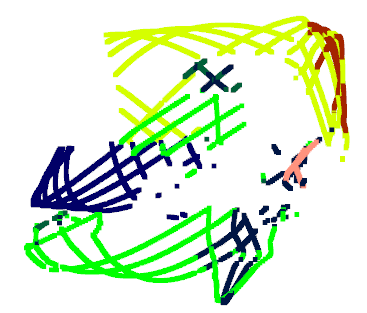}}
{\includegraphics[width=0.24\textwidth]{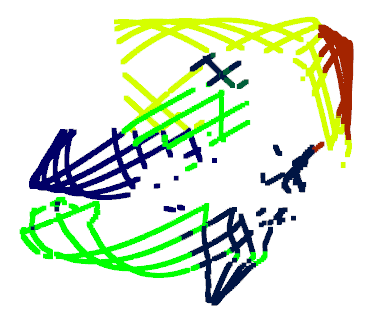}}
% Baseline
% {\includegraphics[width=0.24\textwidth]{figs/visuals/idx_23_scaleFull_input.png}}
% {\includegraphics[width=0.24\textwidth]{figs/visuals/idx_23_scale0_input.png}}
% {\includegraphics[width=0.24\textwidth]{figs/visuals/idx_23_scale1_input.png}}
% {\includegraphics[width=0.24\textwidth]{figs/visuals/idx_23_scale3_input.png}}
% Resscal3D
{\includegraphics[width=0.24\textwidth]{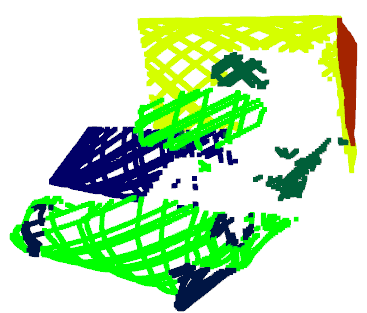}}
{\includegraphics[width=0.24\textwidth]{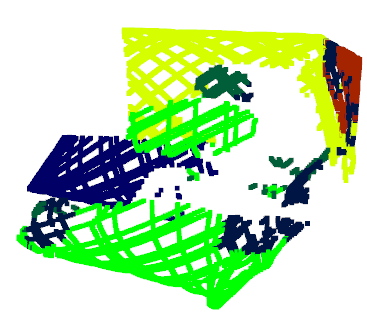}}
{\includegraphics[width=0.24\textwidth]{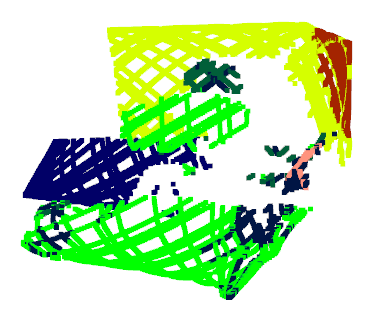}}
{\includegraphics[width=0.24\textwidth]{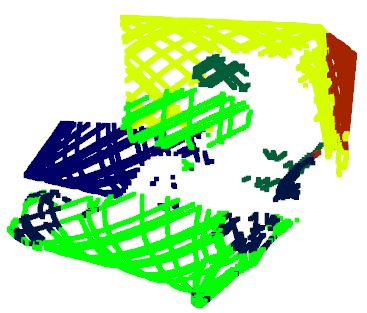}}
% \subcaptionbox*{(d)\\Scale 2\\Time: \textcolor{red}{???} s}{\includegraphics[width=0.19\textwidth]{figs/arch/resscal_fusion_arch.png}} 
% RESSCAL3D++
\subcaptionbox*{(a)\\Ground truth}{\includegraphics[width=0.24\textwidth]{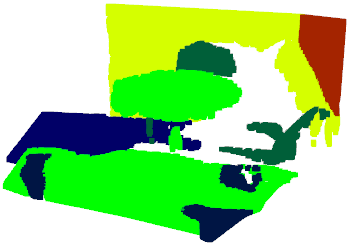}}
\subcaptionbox*{(b)\\Non-scalable baseline~\cite{zhao2021point}}{\includegraphics[width=0.24\textwidth]{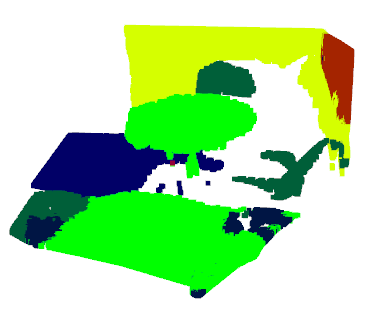}}
\subcaptionbox*{(c)\\RESSCAL3D~\cite{royen2023resscal3d}}{\includegraphics[width=0.24\textwidth]{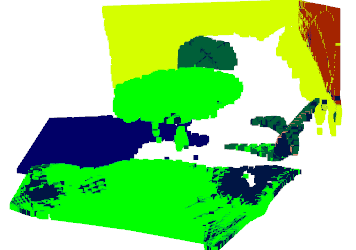}}
\subcaptionbox*{(d)\\RESSCAL3D++}{\includegraphics[width=0.24\textwidth]{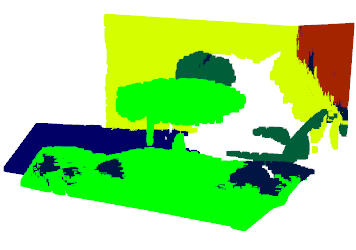}}
\captionsetup{justification=justified}
\caption{Visualization of the non-scalable baseline~\cite{zhao2021point}, RESSCAL3D~\cite{royen2023resscal3d} and RESSCAL3D++ on VX-S3DIS. The semantic predictions after the third, fourth and fifth (last) scale are visualised on the top, middle and bottom row, respectively.}
\label{fig:vis_results_vx_s3dis}
\end{figure*}

% \subsection{Implementation details}
% \label{subsec:impl_details}
\noindent\textbf{Implementation details.} The experiments on S3DIS were conducted by following the settings from \cite{royen2023resscal3d} to allow for a fair comparison. At the exception of the optimizer and learning rate, we also preserved these settings for the experiments on VX-S3DIS. We have empirically opted for the Adam optimizer and a base learning rate of 0.01. We have chosen for 5 scales with partitioning timestamps 2000, 6000, 15000, 35000 and 65536. As backbone architecture for semantic segmentation, we utilized one of the state-of-the-art methods, PointTransformer \cite{zhao2021point}, for consistency and easy comparison with RESSCAL3D. Important to note is that the backbone can be freely chosen. RESSCAL3D++ is trained scale by scale.

% \subsection{S3DIS semantic segmentation}
\noindent\textbf{S3DIS semantic segmentation.} We compare the performance of the proposed method against the scalable and non-scalable state-of-the-art in Figure \ref{fig:s3dis_mIoU_results}. The non-scalable baseline employs the same semantic segmentation backbone, namely \cite{zhao2021point}. While the scalable approaches process only the additional points at each scale, using side information from the previous scales, the baseline processes the whole point cloud at that scale. Thus, the latter can only be launched when all data is available and does not process data in a progressive manner. Also, no intermediate results are obtained. In Figure \ref{fig:s3dis_mIoU_results} can be noticed that RESSCAL3D++ does not only outperform the only other resolution-scalable work, RESSCAL3D, at all scales, but also reduces the cost of scalability at the highest scale from 2.1\% to 0.6\%. In Figure \ref{fig:s3dis_time_results} can be seen that this increase in performance goes along with a negligible effect on the inference latency. The scalable approach shows a clear benefit over the non-scalable baseline since it is not only 31 to 62\% faster at the highest scale, but also enables early predictions, the first one after merely 6\% of the total inference time. We represent the induced latency by RESSCAL3D++ as a zone since it starts the processing during the otherwise lost acquisition time, while the non-scalable baseline can only start once all data is available. Therefore, depending on the speed of acquisition, the latency induced by RESSCAL3D++ will be bounded to the yellow zone.

Qualitative results are presented in Figure \ref{fig:s3dis_visuals}. Note that RESSCAL3D is unable to refine previous predictions which are therefore also present in the final predictions at the highest scales. RESSCAL3D++ on the other hand is able to correct those, leading to a more consistent and correct prediction at the highest scales. This is particulary noticeable for the erronous clutter (black) prediction in the bookcase (pink) on the righthand side. Also noteworthy is that for this visual example, RESSCAL3D++ even outperforms the non-scalable baseline. One reason could be that the processing on different resolution levels allows to increase performance in some cases.

% \subsection{S3DIS semantic segmentation}
\noindent\textbf{VX-S3DIS semantic segmentation} As a second experiment, we apply the proposed method on our novel dataset VX-S3DIS. A similar evaluation strategy as for the S3DIS semantic segmentation is followed. From Figure \ref{fig:vx_miou_results} can be noticed that all RESSCAL3D++ allows an important improvement over RESSCAL3D with the reduction of the cost of scalability from 2.0\% to only 0.2\%. In terms of inference latency, Figure \ref{fig:vx_time_results} shows that, depending on the acquisition time, a reduction in inference time at the highest scale can be reduced by 15.6-63.9\%. Also, the first predictions at the lowest scale can be obtained at only 7\% of the inference time of the baseline at the highest scale.

Qualitative results are presented in Figure \ref{fig:vis_results_vx_s3dis}. Similar to the S3DIS experiments above, there can be noticed that RESSCAL3D solves the inconsistencies of the different predictions. This is particularly noticeable for the chair (dark green) and board (red) at the highest scales. Our proposed method even outperforms the baseline for the table (light green) at the bottom left corner.
\section{Conclusion}
\label{sec:conclusion}

In this paper, we introduced the first point cloud dataset emulating a resolution-scalable 3D sensor. Additionally, we presented a novel method, RESSCAL3D++, which surpasses the current state-of-the-art. By enabling joint data acquisition and processing, our approach achieves significantly reduced processing latencies and enables extremely early predictions.

% \section{Acknowledgement}
% \label{sec:ack}

% This is acknowledgement

% Below is an example of how to insert images. Delete the ``\vspace'' line,
% uncomment the preceding line ``\centerline...'' and replace ``imageX.ps''
% with a suitable PostScript file name.
% -------------------------------------------------------------------------
% \begin{figure}[htb]

% \begin{minipage}[b]{1.0\linewidth}
%   \centering
%   \centerline{\includegraphics[width=8.5cm]{image1}}
% %  \vspace{2.0cm}
%   \centerline{(a) Result 1}\medskip
% \end{minipage}
% %
% \begin{minipage}[b]{.48\linewidth}
%   \centering
%   \centerline{\includegraphics[width=4.0cm]{image3}}
% %  \vspace{1.5cm}
%   \centerline{(b) Results 3}\medskip
% \end{minipage}
% \hfill
% \begin{minipage}[b]{0.48\linewidth}
%   \centering
%   \centerline{\includegraphics[width=4.0cm]{image4}}
% %  \vspace{1.5cm}
%   \centerline{(c) Result 4}\medskip
% \end{minipage}
% %
% \caption{Example of placing a figure with experimental results.}
% \label{fig:res}
% %
% \end{figure}

%\vfill
%\pagebreak

\vfill\pagebreak

% \section{REFERENCES}
% \label{sec:refs}

% List and number all bibliographical references at the end of the
% paper. The references can be numbered in alphabetic order or in
% order of appearance in the document. When referring to them in
% the text, type the corresponding reference number in square
% brackets as shown at the end of this sentence \cite{C2}. An
% additional final page (the fifth page, in most cases) is
% allowed, but must contain only references to the prior
% literature.

\bibliographystyle{IEEEbib.sty}
\bibliography{refs}

\end{document}